\documentclass[letterpaper, 10 pt, conference]{ieeeconf}  % Comment this line out if you need a4paper

\IEEEoverridecommandlockouts                              % This command is only needed if 
\usepackage{pifont}
\usepackage{booktabs}
\usepackage{graphicx}
\usepackage{amsmath} % For \text command
\usepackage{multirow}
\usepackage{float}
\usepackage{subcaption}
\makeatletter
\let\NAT@parse\undefined
\makeatother
\usepackage[colorlinks=true, urlcolor=black, linkcolor=black, citecolor=black]{hyperref}

\title{\LARGE \bf
Glovity: Learning Dexterous Contact-Rich Manipulation via Spatial Wrench Feedback Teleoperation System
}

\let\oldtwocolumn\twocolumn
\renewcommand\twocolumn[1][]{
    \oldtwocolumn[{#1}{
    \begin{center}
        \begin{minipage}{0.6\textwidth}
        \centering
        \includegraphics[width=\textwidth]{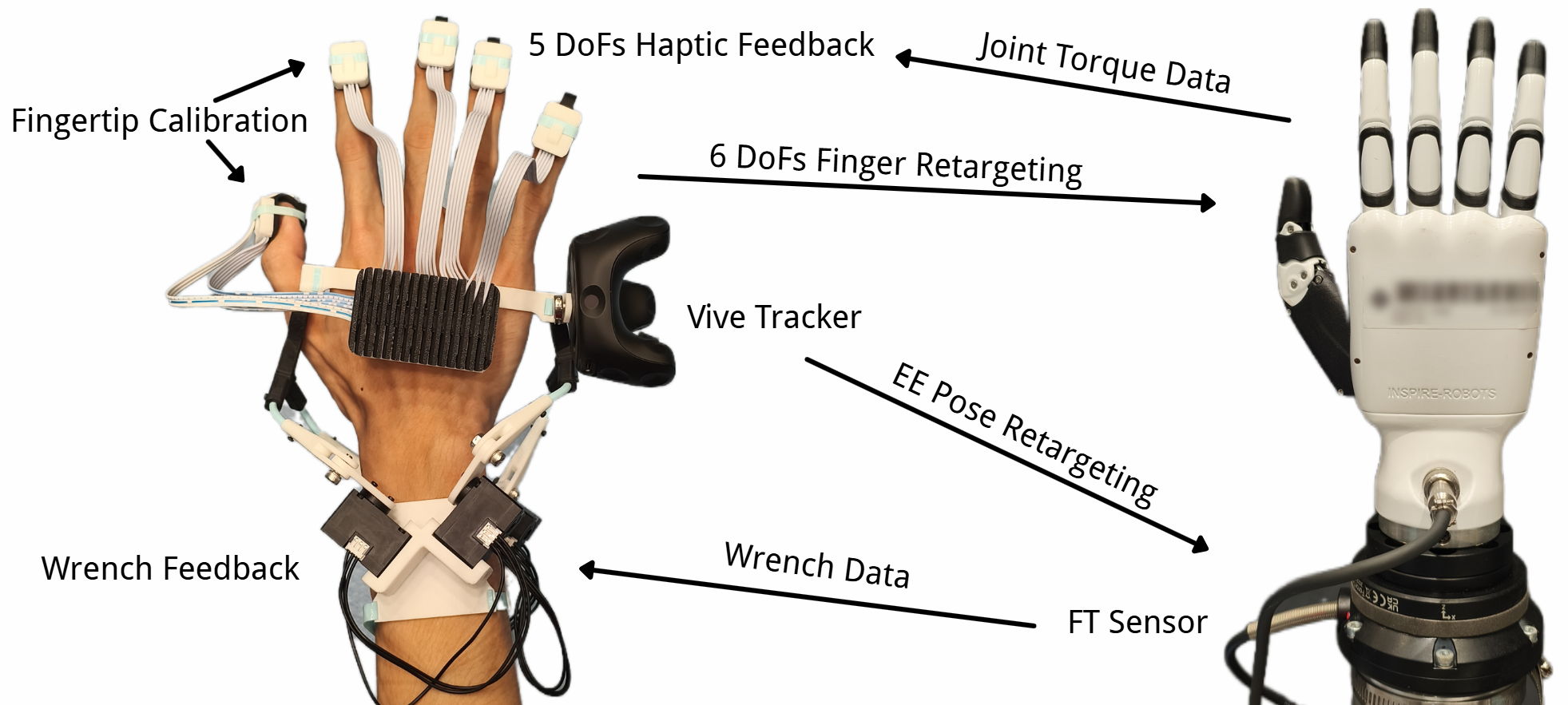} % Placeholder for actual filename
        \subcaption{Glovity Teleoperation System} \label{fig:motion}
    \end{minipage}
    \hfill
    \begin{minipage}{0.32\textwidth}
        \centering
        \vspace{1mm}
        \includegraphics[width=\textwidth]{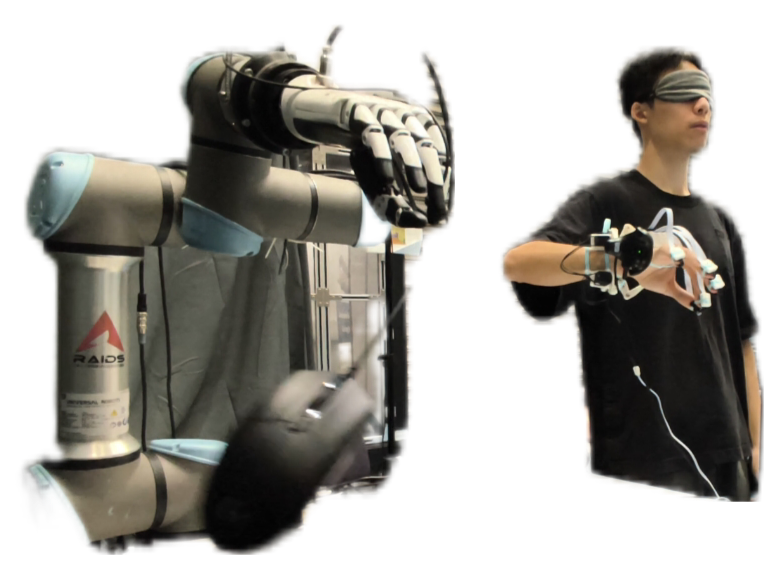} % Placeholder for actual filename
        \vspace{1mm}
        \subcaption{Blindfolded Mouse Swinging} \label{fig:mouse}
    \end{minipage}
   
    \vspace{3mm} % Adjusting spacing between rows
   
    \hspace{10mm}
    \begin{minipage}{0.2\textwidth}
        \centering
        \vspace{2mm}
        \includegraphics[width=\textwidth]{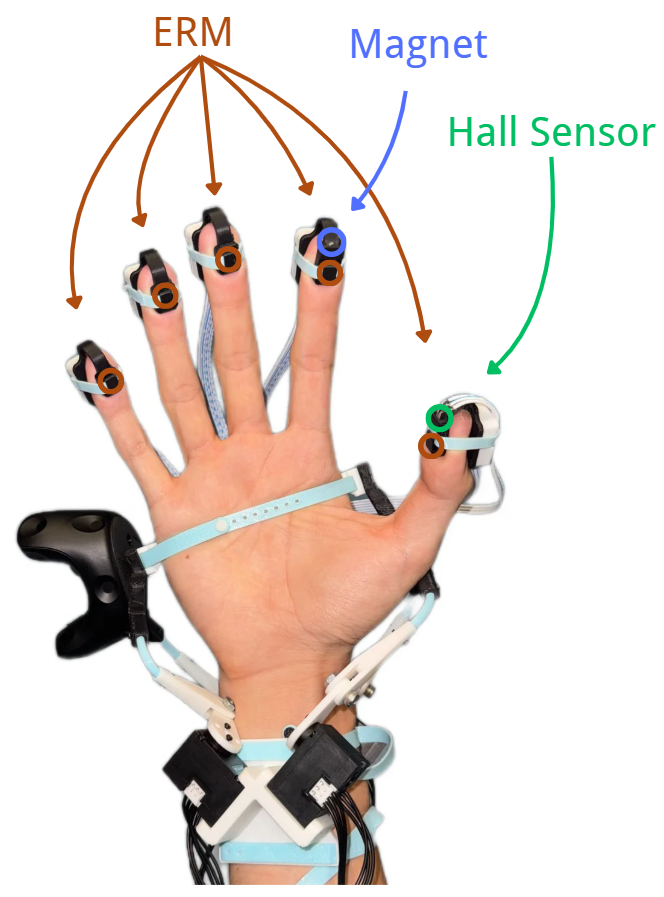} % Placeholder for actual filename
        \subcaption{Haptic Glove with Fingertip Hall Calibration} \label{fig:leap}
    \end{minipage}
    \hfill
    \begin{minipage}{0.32\textwidth}
        \centering
        \vspace{4mm}
        \includegraphics[width=\textwidth]{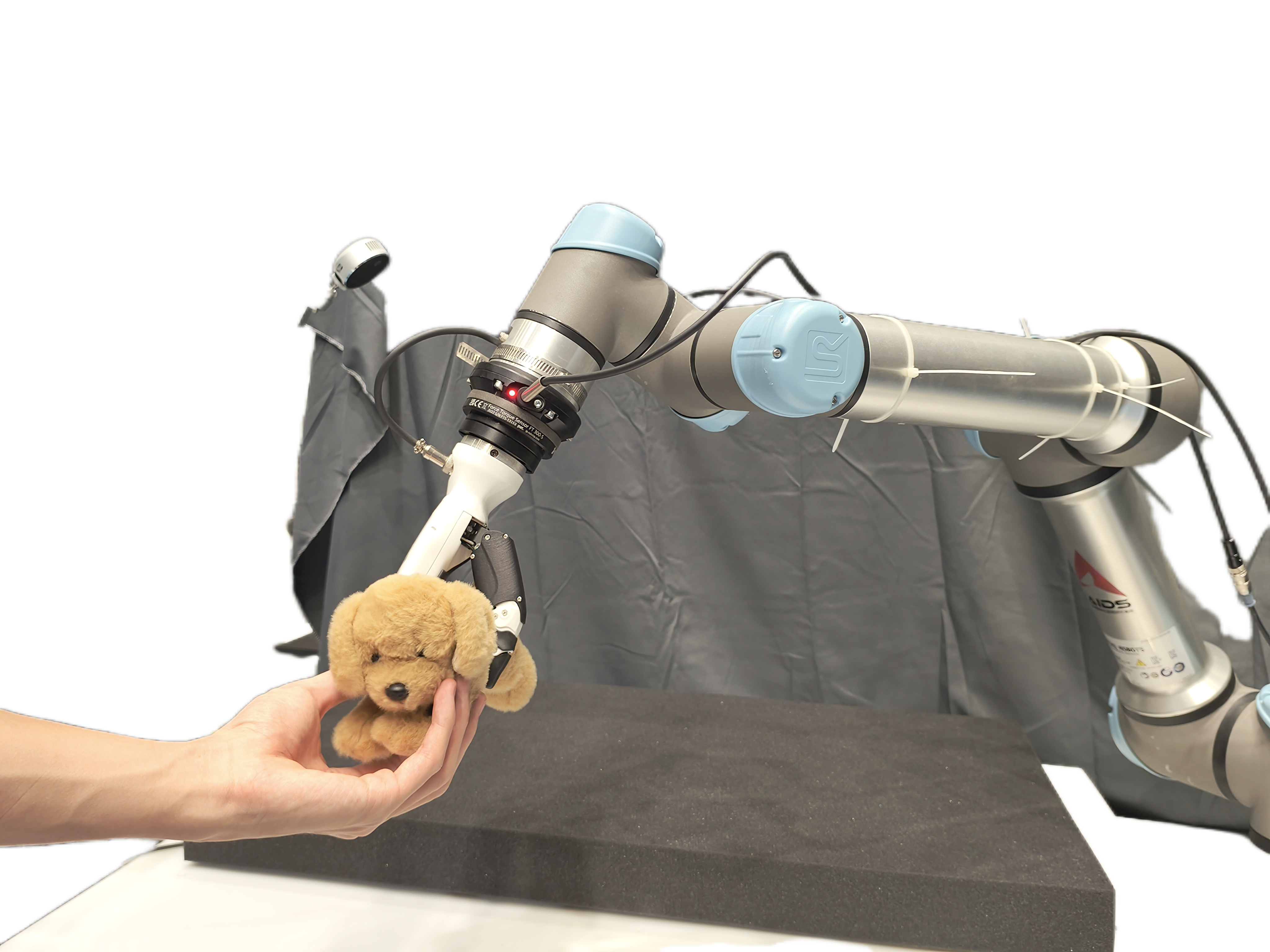} % Placeholder for actual filename
        \subcaption{Handing Toy Outside Camera's Field of View} \label{fig:hand_over}
    \end{minipage}
    \hfill
    \begin{minipage}{0.3\textwidth}
        \centering
        \vspace{9mm}
        \includegraphics[width=\textwidth]{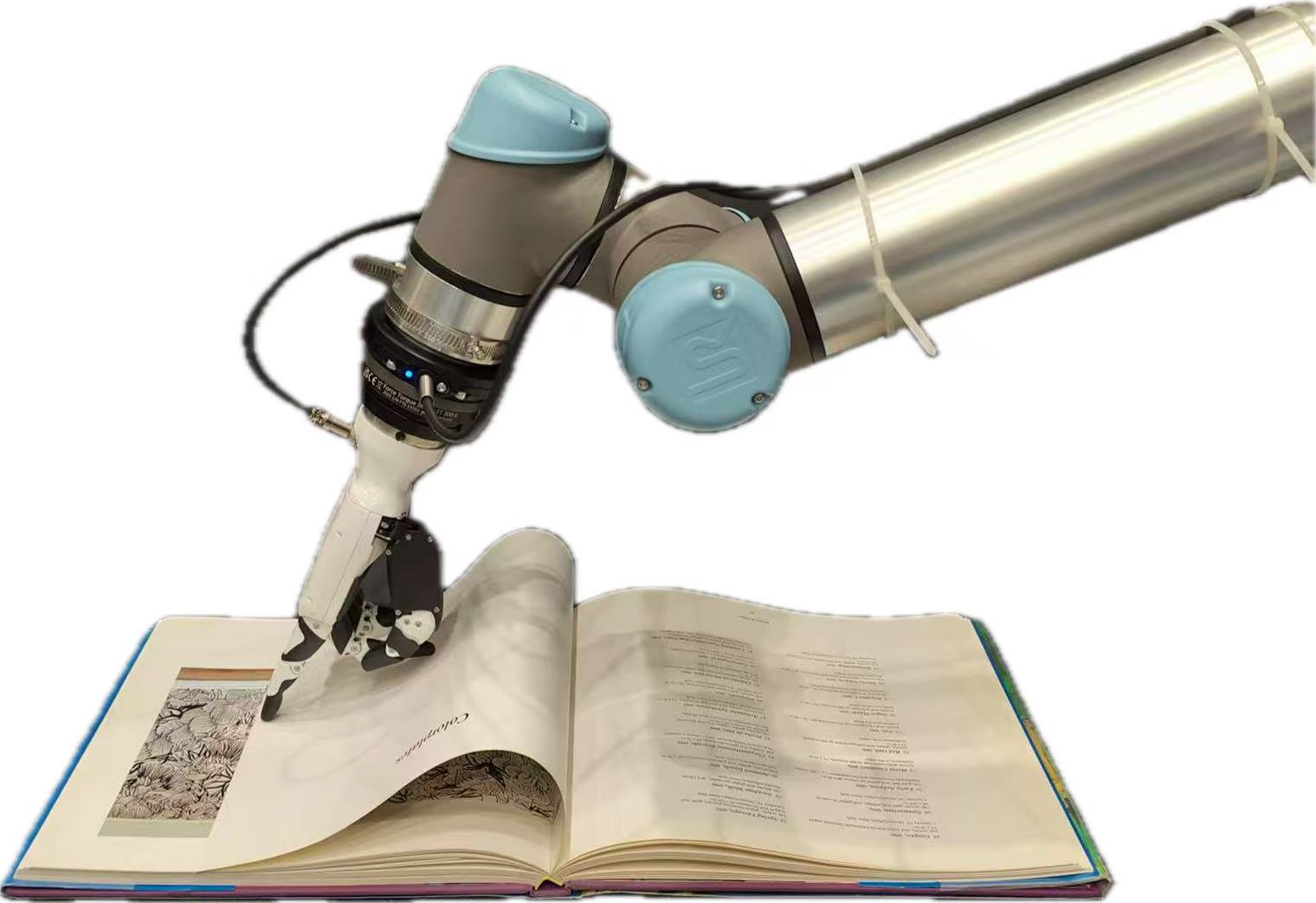} % Placeholder for actual filename
        \subcaption{Page Flipping at Different Heights} \label{fig:haptic}
    \end{minipage}
   
    \vspace{1mm} % Adjusting spacing before caption

\caption{Overview of Glovity. (a) and (c) illustrates the overall system architecture, comprising a wrist-mounted wrench feedback system and a haptic glove, designed to facilitate intuitive force feedback and precise teleoperation. (b) Performance of the wrench feedback in high-dynamic scenarios, enabling the operator to accomplish mouse swinging tasks relying solely on the provided feedback. (d) and (e) showcase two imitation learning tasks, demonstrating the capability of the Glovity system to collect high-quality data and its potential for imitation learning with force.}
\label{fig:main}

    \end{center}
    }]
}

\author{Yuyang Gao, Haofei Ma and Pai Zheng$^{*}$% <-this % stops a space
}
\begin{document}

% \maketitle
\thispagestyle{empty}
\pagestyle{empty}

\begin{figure*}[htbp]

\maketitle    
        
\end{figure*}

%%%%%%%%%%%%%%%%%%%%%%%%%%%%%%%%%%%%%%%%%%%%%%%%%%%%%%%%%%%%%%%%%%%%%%%%%%%%%%%%
\begin{abstract}
We present Glovity, a novel, low-cost wearable teleoperation system that integrates a spatial wrench (force-torque) feedback device with a haptic glove featuring fingertip Hall sensor calibration, enabling feedback-rich dexterous manipulation. Glovity addresses key challenges in contact-rich tasks by providing intuitive wrench and tactile feedback, while overcoming embodiment gaps through precise retargeting. User studies demonstrate significant improvements: wrench feedback boosts success rates in book-flipping tasks from 48\% to 78\% and reduces completion time by 25\%, while fingertip calibration enhances thin-object grasping success significantly compared to commercial glove. Furthermore, incorporating wrench signals into imitation learning (via DP-R3M) achieves high success rate in novel contact-rich scenarios, such as adaptive page flipping and force-aware handovers. All hardware designs, software will be open-sourced. Project website:  \href{https://glovity.github.io/}{glovity.github.io} %to facilitate community advancements in robotic learning. Project Website: \url{https://glovity.github.io}

\end{abstract}

%%%%%%%%%%%%%%%%%%%%%%%%%%%%%%%%%%%%%%%%%%%%%%%%%%%%%%%%%%%%%%%%%%%%%%%%%%%%%%%%

\section{INTRODUCTION}

Robotic teleoperation has emerged as a cornerstone for enabling robots to perform complex, contact-rich manipulation tasks, bridging the gap between human dexterity and robotic precision. Recent advancements in imitation learning have demonstrated the potential of human demonstrations to train robots for tasks requiring intricate interactions. However, existing teleoperation systems face significant challenges that limit their efficacy in such scenarios. First, the lack of multimodal feedback—beyond visual cues—restricts operators' ability to perceive and adjust to real-time force and tactile interactions, which are critical for tasks involving delicate or dynamic contact \cite{babarahmati2021robust,pacchierotti2023cutaneous}. Second, structural disparities between human and robotic manipulators, coupled with retargeting latency, introduce an embodiment gap that complicates precise control~\cite{o2024open}. These limitations result in inefficient data collection and suboptimal performance in dexterous manipulation tasks.

To address these challenges, we introduce Glovity, a novel, low-cost, and open-source teleoperation system designed to enhance contact-rich manipulation through spatial wrench feedback and haptic glove integration. Glovity combines a wearable wrench feedback mechanism, which provides intuitive force and torque cues without restricting arm mobility, with a haptic glove equipped with fingertip Hall calibration for precise grasping. Costing less than \$300 and utilizing 3D-printed components, Glovity is accessible and replicable, requiring only a few hours for assembly. Unlike traditional systems that rely heavily on expensive haptic devices or bulky exoskeletons \cite{buamanee2024bi,peuchpen2024real}, Glovity achieves effective feedback and efficient retargeting through innovative hardware and algorithmic designs, ensuring compatibility with diverse robotic platforms.

The main contribution of this work are as follows:
\begin{itemize}
    \item \textbf{Wearable Spatial Wrench Feedback System}: We develop a compact, wearable system that delivers real-time, multi-dimensional force-torque feedback, enabling intuitive operator adjustments in contact-rich tasks without compromising arm mobility.
    \item \textbf{Haptic Glove with Fingertip Hall Calibration}: We propose a haptic glove integrated with linear Hall sensors for precise fingertip calibration, significantly improving the accuracy of grasping thin objects by mitigating structural misalignments.
    \item \textbf{Efficient Demonstration Data for Imitation Learning}: We demonstrate Glovity’s capability to collect high-quality demonstrations, incorporating wrench signals into diffusion-based imitation learning policies to achieve robust performance in novel, contact-rich tasks.
\end{itemize}

Through comprehensive user studies and imitation learning experiments, we demonstrate Glovity’s ability to provide intuitive and effective feedback for precise control in contact-rich and high-dynamic tasks, as well as its efficient grasping capabilities for thin objects. All hardware and software will be open-sourced, fostering community-driven innovation in robotic learning.

\section{RELATED WORKS}

\subsection{Teleoperation with End-effector Force Feedback}

Currently, the majority of teleoperation systems rely solely on visual feedback, including leader-follower arm setups~\cite{wu2023gello,zhao2023learningfinegrainedbimanualmanipulation}, VR-based interfaces~\cite{cheng2024opentelevisionteleoperationimmersiveactive,arunachalam2022holodexteachingdexterityimmersive}, and camera-based tracking~\cite{qin2024anyteleopgeneralvisionbaseddexterous,song2020graspingwildlearning6dofclosedloop}. This exclusive dependence on visual cues renders these systems inadequate for precise contact-rich tasks. Some works have improved upon leader-follower arm systems by incorporating bilateral force feedback through methods like shared joint torque~\cite{buamanee2024bi,wu2025robocopilothumanintheloopinteractiveimitation,bazhenov2025echoopensourcelowcostteleoperation,liu2025factr}. However, these systems are typically tailored to specific robots and, due to structural constraints, are challenging to adapt for dexterous manipulation. Other approaches utilize haptic devices to enable force feedback teleoperation~\cite{he2025foar,peuchpen2024real,meli2017experimental}, but such devices are typically expensive and have limited operation space, making them difficult to apply to a wide range of tasks.
In contrast, Glovity's wrench feedback system achieves low-cost implementation while being wearable on the human wrist and compatible with dexterous operations, providing operators with highly enriched feedback.
\subsection{Dexterous Hand Teleoperation}
Efficient teleoperation of dexterous hands has long been a challenge in robot learning, primarily due to structural differences between robotic and human hands, which necessitate effective mapping algorithms to bridge the gap. Several approaches have been developed to address this. Vision-based systems, such as AnyTeleop \cite{qin2024anyteleopgeneralvisionbaseddexterous}, achieve effective hand mapping through a flexible framework supporting diverse robotic arms, hands, and camera setups. Bunny-VisionPro \cite{ding2024bunnyvisionprorealtimebimanualdexterous} leverages VR devices and haptic feedback to provide real-time bimanual dexterous control, enhancing immersion for imitation learning. However, vision-based methods often suffer from occlusion issues and increased latency due to real-time pose estimation.
Alternatively, some works integrated commercial gloves~\cite{xu2025immersivevirtualrealitybimanual,shaw2024bimanualdexteritycomplextasks,lu2024teleoperated}, and there are also Low-cost alternatives, such as DOGlove~\cite{zhang2025doglovedexterousmanipulationlowcost}, employ an exoskeleton-encoder scheme with motion capture system and cable-driven torque feedback, achieving precise hand teleoperation and fingertip positioning. Yet, such exoskeleton-based designs can be bulky, compromising user comfort.
In contrast, Glovity utilizes an IMU+Hall sensor approach, combining minimal components with innovative structural and algorithmic designs to achieve efficient hand mapping and precise fingertip calibration. This wearable, low-latency solution avoids occlusion issues and bulky hardware, offering enhanced performance for contact-rich dexterous tasks while maintaining low-cost and adaptability.
\begin{figure*}[h]
    \centering
    \begin{minipage}{\textwidth}
        \centering
        \includegraphics[width=\textwidth]{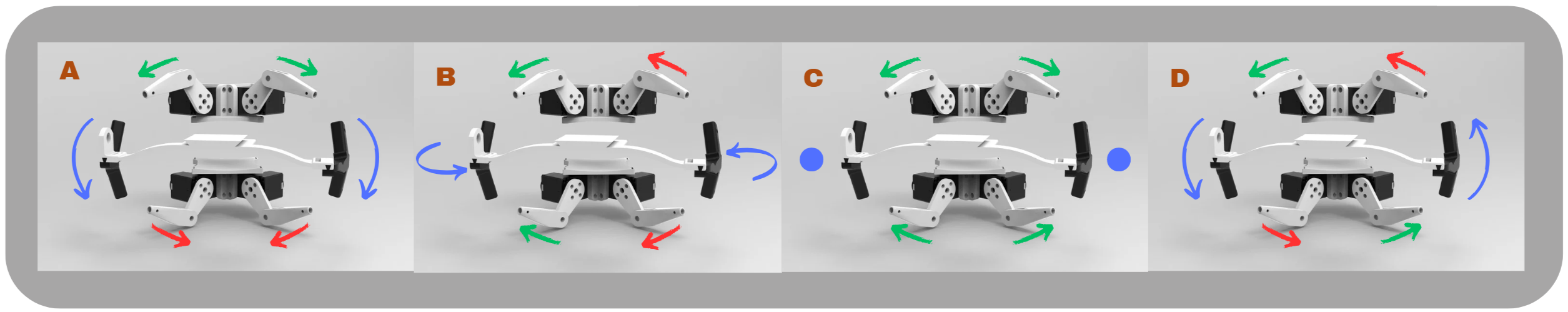} % Replace with your image file
        \caption{Illustration of the structure and four base feedback modes of the wrench feedback system in response to base components. (A) Represents the force applied along the x-axis, inducing a Vertical motion pattern. (B) Depicts the force along the y-axis, driving a horizontal motion. (C) Indicates the force along the z-axis, eliciting a forward-backward motion. (D) Shows the torque around the z-axis, resulting in a rotational response.}
        \label{fig:setup}
    \end{minipage}
\end{figure*}
\subsection{Imitation learning with Force/Tactile Data}

Research on imitation learning has been around for many years, originating from foundational works in behavioral cloning and apprenticeship learning in the early 2000s. As the tasks assigned by humans to robots become progressively more complex and involve richer contact interactions and longer time horizons, integrating force/tactile signals into imitation learning algorithms has become an irreversible trend~\cite{zhu2025touch,liu2024forcemimic,xue2025reactive,liu2025factr,kim2024goal}. Among these, RDP~\cite{xue2025reactive} incorporates force/tactile as conditional inputs while employing a fast-slow architecture to fine-tune actions, delivering impressive results in contact-rich tasks. ForceMimic~\cite{liu2024forcemimic} uses a handheld gripper for data collection, predicts position-force pairs, and uses hybrid position-force control, excelling in the peeling task. The above methods show great potential for incorporating force/tactile sensing into imitation learning, but unfortunately they are not compatible with dexterous manipulation. In this project, we incorporate wrench data as observations into the DP-R3M model, exploring novel scenarios and further demonstrating the potential of multimodal imitation learning in dexterous manipulation.

\section{TELEOPERATION SYSTEM DESIGN}

\subsection{Design Philosophy}Glovity is designed to provide rich, intuitive feedback along with excellent retargeting performance for dexterous manipulation. It significantly enhances the efficiency and quality of data collection while ensuring the entire system is low-cost, easy to replicate and assembly.
Glovity's design adheres to the following principles:
\begin{enumerate}
\item \textbf{High Cost-Performance}: Glovity's core goal is to be affordable and highly effective. By using a minimal number of commercial high-quality servos and sensors, combined with 3D-printed parts, we keep the total cost around \$300 (\$230 for wrench feedback and \$70 for haptic glove). Through ingenious structural and algorithmic designs, it achieves rich feedback, efficient retargeting and calibration, delivering impressive performance in real-world use.
\item \textbf{Ease of Replication and Assembly}: All electronic and structural components of Glovity can be easily obtained online and via 3D printing. Additionally, it adopts a Lego-style component design, allowing the entire assembly process to be completed in just a few hours with a few screws.
\item \textbf{Versatility and modularity}: Glovity fits hands of various sizes and suitable for different robots and dexterous hands, offering comfortable wear and low latency (up to 100Hz for wrench feedback, retargeting, and tactile feedback) to handle diverse manipulation tasks. Glovity also adopts a modular design, allowing the wrench feedback and the tactile glove to be used separately, making it adaptable to various scenarios.
\end{enumerate}
In summary, through these principles, Glovity offers an efficient and economical solution for dexterous operations, suitable for robotics researchers and developers.

\subsection{Wrench Feedback System}
Glovity’s wrench feedback system is designed to provide intuitive wrench feedback in a compact, ergonomic design. For practical considerations, the system is designed to provide feedback in \textbf{four primary dimensions}, with forces and torques along the x and y axes (where the x-axis passes through the palm, while the y-axis represents the axis of finger flexion) implemented using a fused feedback approach (detailed in later section). This design accounts for the fact that, in most manipulation tasks, forces and torques at the robot's end-effector along the x and y axes tend to occur simultaneously and in the same direction.
\subsubsection{Hardware and Mechanism Design}The system employs four XL330 servomotors, selected for their high torque and rapid response, coupled with a dual-linkage mechanism to convert rotational motion into linear displacement and rotation of a palm-mounted fixator, this fixator is connected via a TPU-printed linkage, which minimizes restrictions on wrist movement while maintaining effective force feedback. The main body of the system is 3D-printed using PLA for structural rigidity, complemented by a TPU strap to securely fasten the device to the user’s forearm. 

The servomotors are arranged in pairs above and below the distal end of the forearm, with adjacent motor pairs oriented at a 90-degree angle, as illustrated in Figure~\ref{fig:setup}. This configuration is driven by two key design considerations:

\begin{itemize}
    \item \textbf{Compact Multi-Directional Feedback}: The orthogonal servomotor arrangement enables the encoding of forces and torques from multiple directions within a compact form factor, enhancing the realism of haptic feedback.
    \item \textbf{Ergonomic Optimization}: Positioning servomotors above and below aligns with the elliptical shape of the human forearm, reducing pressure points and improving wearability.
\end{itemize}
\subsubsection{Wrench Feedback Mapping}
The primary principle of the wrench feedback is to control the rotation of servomotors to corresponding positions based on real-time 6D force-torque data acquired from the FT sensor mounted at the robot's end-effector.

For each servomotor \( i \in \{1, 2, 3, 4\} \), the mapping algorithm can be expressed as:
\begin{equation}
\theta_i = \theta_{\text{init}_i} + \sum_{j \in \{f_y, f_x, f_z, \tau_z\}} w_j \cdot \kappa_j \cdot s_{j_i} \cdot c_{j_i} \cdot j \label{eq:1}
\end{equation}
where \(\theta_{\text{init}_i}\) is the initial/no-load angle for servomotor \(i\); \(w_j\) is the weight of component \(j\), based on its proportion among all components; \(\kappa_j\) is the sensitivity of component \(j\), manually set based on task requirements; \(s_{j_i}\) is the sign of component \(j\) for servomotor \(i\), determined by the placement of the servo; \(j\) represents the force/torque component, where \(f_y\), \(f_x\), \(f_z\) represent forces (N) and \(\tau_z\) represents torque (Nm); and \(c_{j_i}\) is the dynamic coefficient for component \(j\) on servomotor \(i\), defined as:
\begin{equation}
c(r) = \max(c_{\text{min}}, 1 + \sigma \cdot \kappa_r \cdot (r + \delta)) \label{eq:2}
\end{equation}
with \(r \in \{ \tau_x / f_y, \tau_y / f_x \}\) being the torque-force ratio; \(\kappa_r\) the sensitivity of the ratio, manually set based on task requirements; \(\sigma\) the sign factor (\(\pm 1\)), determining the direction of adjustment based on the ratio type; \(\delta\) the offset parameter, adjusting the pivot point of the ratio’s effect, typically set based on system geometry (e.g., distance from hand center to wrist); and \(c_{\text{min}}\) the minimum coefficient value, ensuring stability by preventing excessive attenuation.
Equation~\eqref{eq:1} defines a servo encoding method that maps the base wrench components $f_x$, $f_y$, $f_z$, and $\tau_z$ to each servomotor’s rotation. As shown in Fig.~\ref{fig:setup}, each component contributes to the servo angle through a specific combination of rotation direction, with their effects summed in proportion to produce the final angle increment. The dynamic coefficient in~\eqref{eq:2} further modulates the contributions of $f_y$ and $f_x$ based on the torque-to-force ratios $\tau_x / f_y$ and $\tau_y / f_x$, respectively, as defined in the second equation. This adjustment primarily aims to simulate sensory differences caused by contact points farther from or closer to the wrist through asymmetric rotations. Additionally, the algorithm employs dynamic time-window filtering to enhance stability, ensuring sensitive responses to sudden changes and stable responses under sustained forces.

\subsection{Haptic Glove}
The IMU-based haptic glove is designed to deliver a cost-effective and user-friendly solution for dexterous teleoperation, enabling immersive experiences through low-latency finger motion capture, precise fingertip calibration, and intuitive haptic feedback. We elaborate on this system across three key aspects: hardware design, retargeting and calibration algorithms, and haptic feedback mechanisms.
\subsubsection{Hardware Design}The Glovity haptic glove employs a compact, modular assembly akin to LEGO blocks, eliminating the need for screws or adhesives during assembly. Drawing on ergonomic principles, it integrates PLA-printed rigid structural components with flexible TPU straps, offering comfort comparable to commercial-grade gloves. The core hardware comprises six WitMotion JY61P IMU modules (five positioned at the fingertips and one on the back of the hand), a 49E Hall sensor (mounted on the thumb pad, paired with a magnet on the index finger), five 3V ERMs (8$\times$2 mm) with L9110s drivers, an XIAO ESP32-S3 microcontroller, and a custom circuit board. All components are integrated within the glove, resulting in a sleek and cohesive design.
\subsubsection{Retargeting and Calibration Algorithms}The retargeting process utilizes the differences in angular velocities and accelerations between the fingertip IMUs and the IMU on the back of the hand to compute finger bend angles. For the index to little fingers, a hybrid acceleration-angular velocity fusion algorithm ensures robustness across diverse postures. The fused angle \(k\) is computed using a complementary filter that blends gyroscope integration with accelerometer-derived angles:
\begin{equation}
\theta_k = \alpha \cdot (\theta_k + \Delta \theta_{\text{gyro}}) + (1 - \alpha) \cdot \theta_{\text{acc}}
\end{equation}
where:
 \(\theta_k\) is the previous fused angle,
 \(\Delta \theta_{\text{gyro}}\) is the gyroscope increment (with corrections for cross-axis influences and damping),
 \(\theta_{\text{acc}}\) is the angle derived from accelerometer projections,
 \(\alpha\) is an adaptive blending factor, to prioritize gyro when acceleration is not reliable.

This fusion ensures robustness against gyroscope drift and accelerometer noise, with additional smoothing via a moving average window of size 6. For the thumb, relative angular velocity is employed, supplemented by Hall sensor-assisted calibration. This leverages sensor readings to quantify the proximity distance between thumb and index fingertips: the sensor is mounted on the thumb pad and paired with a neodymium magnet on the index finger, causing the output value (\(h\)) to increase as magnetic field strength grows with approaching fingertips. Based on sensor sensitivity and practical needs, two key thresholds are predefined: a lower threshold (e.g., \(h_{\text{low}}\)) to detect proximity and trigger calibration when \(h \geq h_{\text{low}}\), and a higher threshold (e.g., \(h_{\text{high}}\)) to indicate fingertip contact when \(h \geq h_{\text{high}}\), setting specific angles per the robotic hand's structure for pinching the fingers together. Within the interval (\(h_{\text{low}} \leq h < h_{\text{high}}\)), transitional angles are computed via linear interpolation and smoothed with low-pass filtering, ensuring the dexterous hand's fingertip pose matches the operator's gesture. This threshold-based method uses empirical sensor calibration to adapt to varying user hand shapes and environmental noise, enabling efficient grasping of thin objects.

\subsubsection{Haptic Feedback Mechanisms}The haptic feedback is derived from force sensors on the dexterous hand, with sensor readings linearly mapped to ERM motor speeds via PWM control, which similar to previous works like~\cite{ding2024bunnyvisionprorealtimebimanualdexterous}. This feedback enhances the teleoperation's immersion and precision, facilitating intuitive control in complex tasks.

\subsection{Hand pose tracking}
Glovity is designed to be compatible with the Vive Tracker for precise hand gesture tracking, with dedicated mounting holes integrated into the haptic glove for seamless attachment. By leveraging the tracker's high-precision positional data, we employ inverse kinematics (IK) to compute joint angles, coupled with servo control, to achieve low-latency retargeting of hand poses to the robotic dexterous hand.

\section{EXPERIMENTS}
By conducting experiments using Glovity, we hope to find answers to the following questions:
\begin{itemize}
    
\item Does the wrench feedback system provide accurate and intuitive feedback while improving the efficiency of teleoperation for contact-rich tasks?
\item How does the haptic glove perform in object grasping tasks, especially thin objects?
\item What is the potential of Glovity for imitation learning, especially for learning with force and torque?
\end{itemize}
To answer the above questions, several experiments were designed and all experiments were conducted using a UR5e robotic arm equipped with an FT300s force/torque sensor and an Inspire Hand RH56. For the user studies, we recruited 10 participants (5 males, 5 females) from the university, none of whom had prior professional teleoperation training. 
\begin{figure}[h]
    \centering
    \begin{minipage}{0.38\textwidth}
        \centering
        \includegraphics[width=\textwidth]{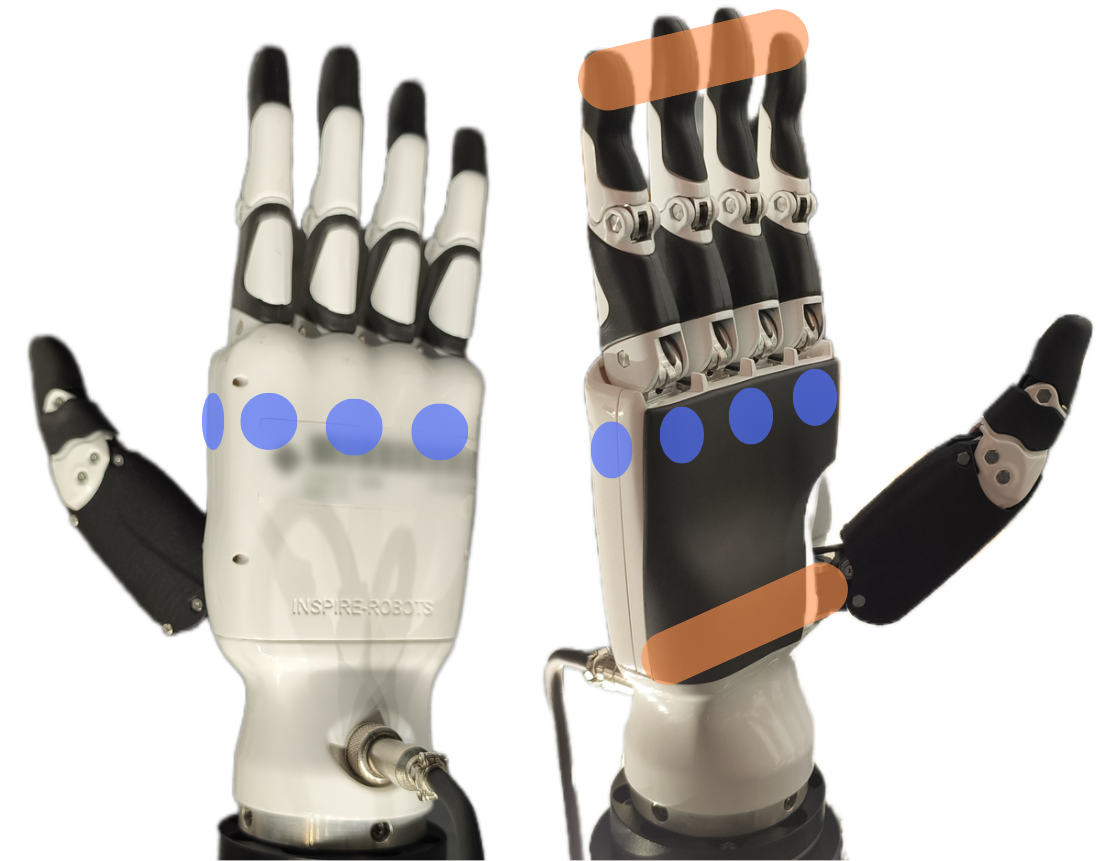} % Replace with your image file
        \caption{The force application regions for the perception experiment. The blue regions correspond to Level 2, comprising eight distinct areas, while the orange regions represent Level 3, consisting of two areas.}
        \label{fig:perception}
    \end{minipage}
\end{figure}
\subsection{User Study on Wrench Feedback}
To address the questions outlined above, we designed three sub-experiments: a perception to evaluate the accuracy and intuitiveness of the feedback, a case study to explore the potential of glovity in highly dynamic tasks, and a teleoperation experiment to assess efficiency improvements in real-world operations.

\subsubsection{Perception Experiment}
In the perception experiment setup, participants wore the wrench feedback system and completed a 3-minute familiarization period to become accustomed to the device. Subsequently, they performed three level perception tasks, ordered from least to most challenging. First, forces were applied along the x, y, or z axis of the FT sensor, and participants identified the force direction. Second, forces were applied to one of 8 distinct areas on the middle section of the Inspire Hand, as shown in Fig~\ref{fig:perception}, and participants identified the specific area. Finally, forces were applied to regions near the fingertips or wrist on the palm, and participants distinguished between these two locations. Each task was repeated 10 times without visual input, with \textbf{success rate} as the evaluation metric, recognition errors or inability to identify as failure.

\begin{figure}[h]
    \centering
    \begin{minipage}{0.45\textwidth}
        \centering
        \includegraphics[width=\textwidth]{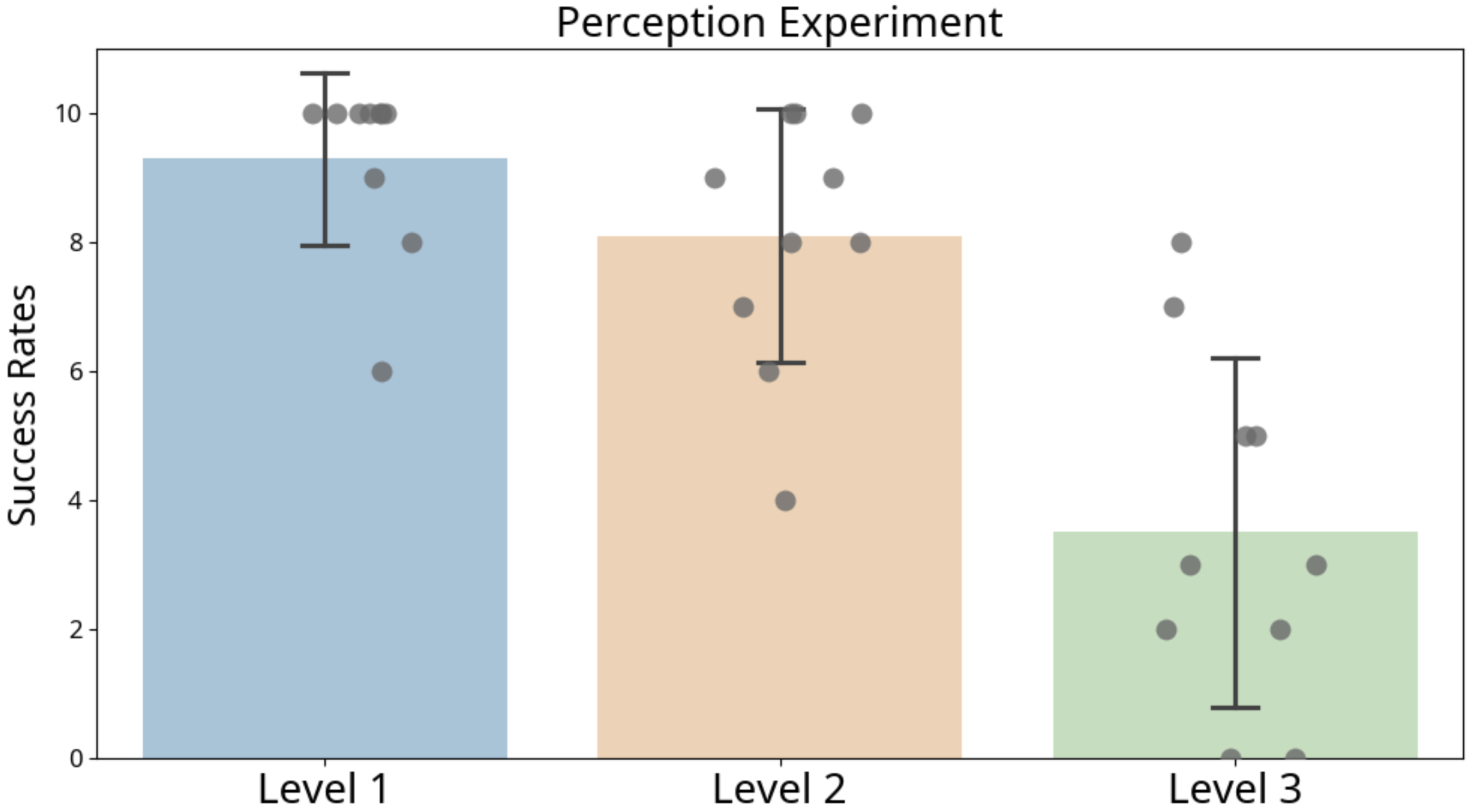} % Replace with your image file
        \caption{Success rates across three experiment levels, where the average success rates of level 1, level 2, and level 3 is 93\%, 81\%, and35\%}
        \label{fig:perceptiontable}
    \end{minipage}
\end{figure}

\textbf{Results:} As shown in Fig~\ref{fig:perceptiontable}, participants in the first level of the perception experiment demonstrated high accuracy in distinguishing forces applied along the x, y, and z axes. In the second level, most participants successfully identified force application areas among the eight designated regions on the middle section of the Inspire Hand, though some showed reduced sensitivity to specific regions, likely due to a mismatch between the device’s size and participants’ hand dimensions, causing slight looseness. In the third level, only a few participants accurately differentiated between force applications near the fingertips and wrist on the palm. This lower performance is attributed to the wrench feedback system’s fixed mechanism at the palm’s middle section, which, despite significant servo rotation differences, provided insufficiently intuitive feedback, hindering participants’ ability to precisely identify the force application region.

\subsubsection{Teleoperation Experiment}
In this experiment, participants performed a page-flipping task five times using the Glovity system, under two conditions: with and without wrench feedback. Flipping thin deformable objects poses challenges in maintaining stable contact and adapting to surface deformation without precise control~\cite{zhao2025learning}. The task's difficulty arose from the need to apply consistent downward force with the index finger, leveraging friction to lift and flip the page successfully. To minimize order effects, five participants initiated the task with the wrench feedback condition, while the remaining began without it. Haptic feedback remained active throughout the experiment; however, its utility was limited due to the near-vertical orientation of the index finger. Performance was evaluated based on two metrics: \textbf{task completion time and success rate}. A trial was deemed successful if the page was flipped in a single attempt without slipping, while failures were recorded if the page slipped mid-process or if excessive force caused the robotic system to stop.
\begin{figure}[h]
    \centering
    \begin{minipage}{0.2\textwidth}
        \hspace{1mm}
        \includegraphics[width=\textwidth]{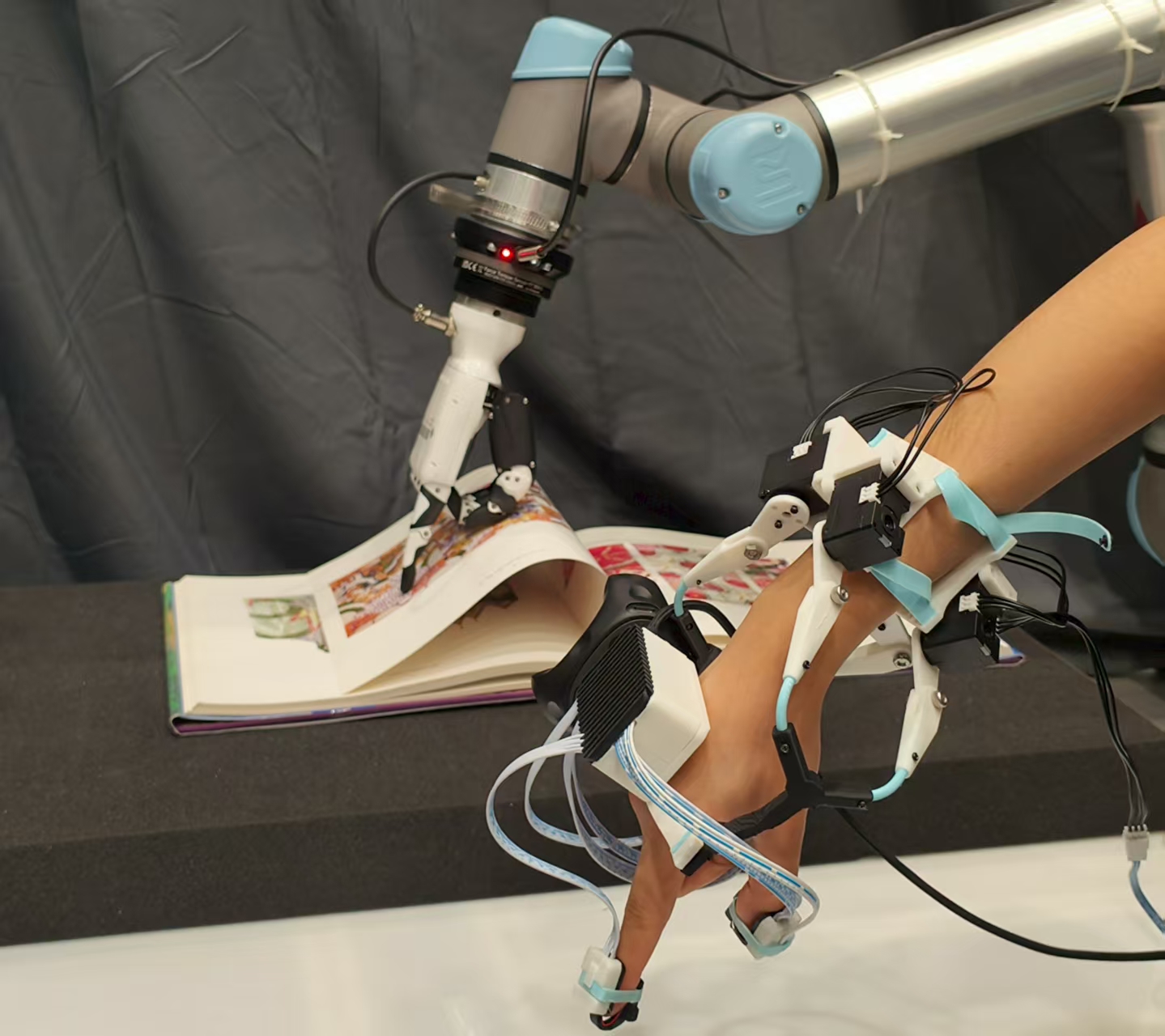} % Replace with your first image file
        \subcaption{}
    \end{minipage}
    \hspace{5mm}
    \begin{minipage}{0.2\textwidth}
        \includegraphics[width=\textwidth]{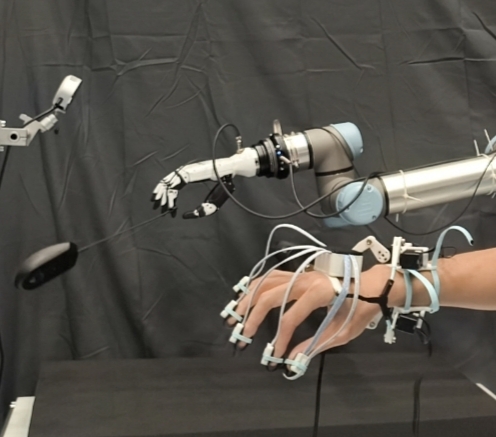} % Replace with your second image file
        \subcaption{}
        \label{fig:mouse2}
    \end{minipage}
    \caption{(a) Teleoperation Experiment on Page Flipping. (b) Case Study on Dynamic Mouse Swinging}
\end{figure}

\textbf{Results:} As shown in Table~\ref{tab:teleop_results}, the experiment results indicate that wrench feedback significantly enhanced the success rate of the book-flipping task while also reducing completion time. The absence of feedback led to a more pronounced increase in instances of excessive force, a trend supported by user reports highlighting the difficulty in accurately assessing contact force magnitude through visual cues alone. Without feedback, operators struggled to perceive whether their fingertip was adequately pressing on the page, often resulting in overcompensation. Wrench feedback effectively addressed this by providing clear sensory cues about fingertip-book contact and pressure levels, enabling users to adjust their force application more precisely and maintain better control throughout the task.
\begin{table}[h]
\centering
\small % Reduce font size to fit within column width
\begin{tabular}{lccc}
\toprule
Wrench Feedback & & \ding{51} & \ding{55} \\
\midrule
Success Rate & & 39/50 & 24/50 \\
Mean Task Time (s) & & 11.8 & 17.7 \\
Failures (Slipping) & & 7 & 15 \\
Failures (Excessive Force) & & 4 & 11 \\
\bottomrule
\end{tabular}
\caption{Teleoperation Experiment Results}
\label{tab:teleop_results}
\smallskip
\end{table}
\subsubsection{Case Study of Dynamic Mouse Swinging}
To investigate the capability of the wrench feedback system in facilitating dynamic tasks, we conducted a case study involving a swinging motion task. A broken computer mouse was selected for its suitable weight and asymmetric mass distribution (heavier at the end). The mouse's cable was securely attached to the index finger of the Inspire Hand. As shown in Fig~\ref{fig:mouse} and~\ref{fig:mouse2}, Operators performed the task via teleoperation, swinging the mouse in a circular motion \textbf{without visual feedback}, relying solely on wrench feedback.

\textbf{Results:} Operators could successfully swing the mouse and modulate both the radius and speed of its motion. This highlights the wrench feedback system's potential to support high-dynamic tasks, enabling effective control in scenarios where visual input is absent.
\begin{figure}[h]
    \centering
    \begin{minipage}{0.19\textwidth}
        \hspace{2mm}
        \includegraphics[width=0.96\textwidth]{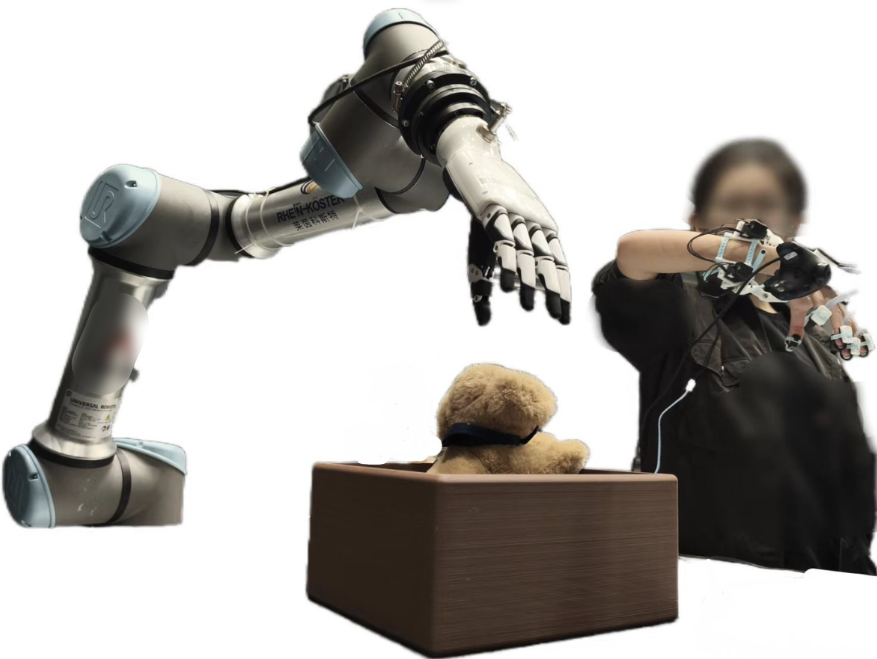} % Replace with your first image file
        \subcaption{Toy Grasping}
    \end{minipage}
    \hspace{2mm}
    \begin{minipage}{0.18\textwidth}
        \includegraphics[width=0.9\textwidth]{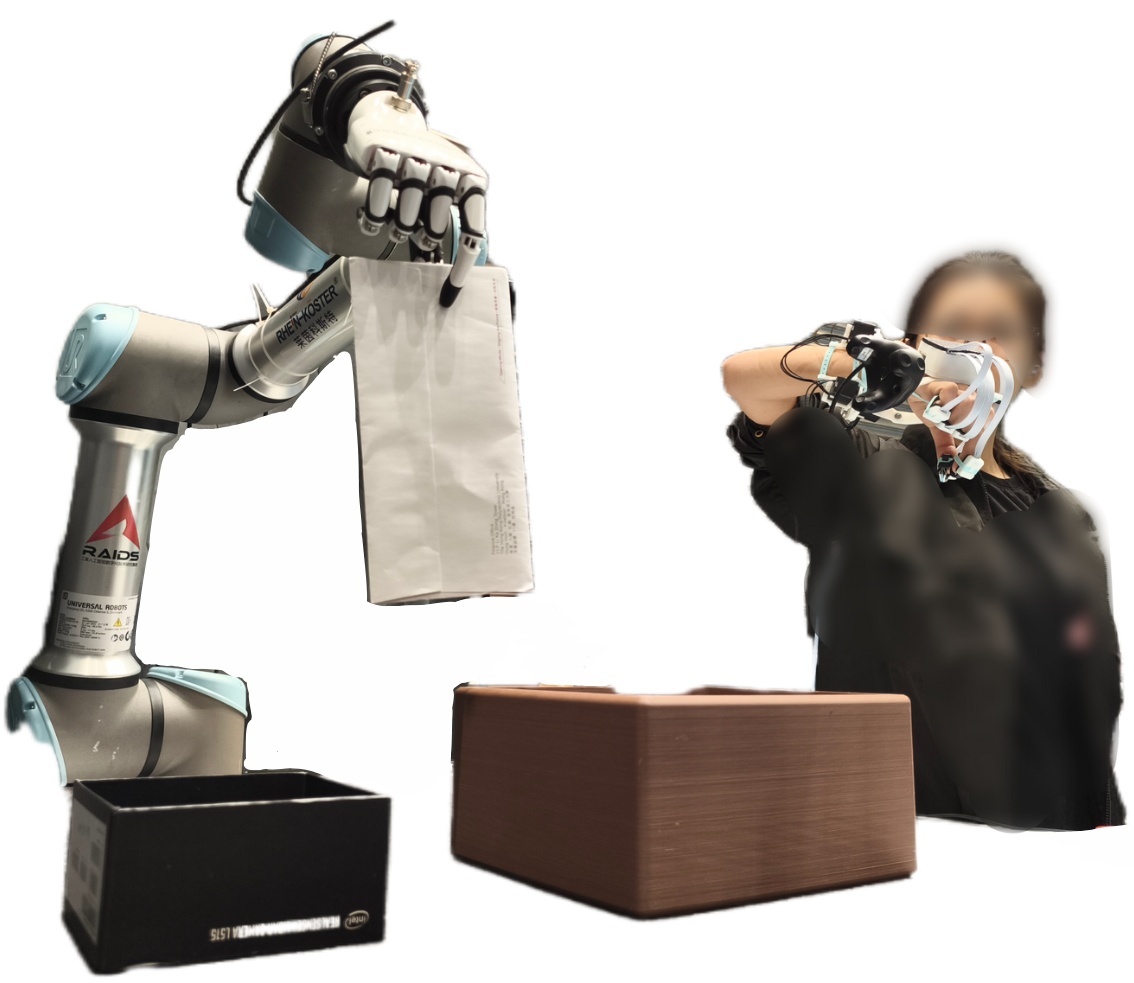} % Replace with your second image file
        \subcaption{Envelop Grasping}
        \label{fig:images}
    \end{minipage}
    \caption{User study on haptic glove}
\end{figure}
\subsection{User Study on Haptic Glove}
To evaluate the grasping capabilities of the haptic glove, participants performed two tasks: picking up a toy dog and an envelope, and placing each into a box, with each task repeated five times. The experiment used AnyTeleop~\cite{qin2024anyteleopgeneralvisionbaseddexterous} and Vrtrix glove pro 11~\cite{VRTRIX2025} as benchmarks, with the robotic arm's end-effector uniformly controlled via a Vive Tracker. The performance metrics includes the success rate and the average time to complete the grasp.

\textbf{Results:} The experimental results demonstrate that, in the toy dog grasping task, the three methods—Glovity, AnyTeleop, and Vrtrix Glove—exhibited comparable performance, characterized by high success rates and short completion times. However, significant differences emerged in the envelope grasping task. AnyTeleop achieved the highest success rate, yet its average completion time was approximately 4 seconds longer than our method. This delay arose from AnyTeleop’s latency and inconsistent detection of thumb-index finger pinching, necessitating users to adjust their hand gestures and retry the pinching action, exacerbated by the absence of haptic feedback. In contrast, the Vrtrix Glove showed a marked decline in success rate and increased completion time, primarily due to its inability to accurately locate fingertip pinching, requiring users to repeatedly adjust thumb and index finger positions to ensure that the Inspire Hand successfully grasped the envelope.
\begin{table}[h]
\centering
\small % Reduce font size to fit within column width
\resizebox{\dimexpr\columnwidth}{!}{
\begin{tabular}{lcccc}
\toprule
Method & \multicolumn{2}{c}{Mean Completion Time (s)} & \multicolumn{2}{c}{Success Rate} \\
\cmidrule(lr){2-3} \cmidrule(lr){4-5}
 & Toy Dog & Envelope & Toy Dog & Envelope \\
\midrule
Ours & 5.2 & 5.9 & 46/50 & 41/50 \\
AnyTeleop & 5.8 & 9.3 & 48/50 & 44/50 \\
Vrtrix Glove & 5.3 & 18.4 & 45/50 & 17/50 \\
\bottomrule
\end{tabular}
}
\caption{Grasping Experiment Results}
\label{tab:grasp_results}
\end{table}
\begin{figure*}[h]
    \centering
    \begin{minipage}{\textwidth}
        \centering
        \includegraphics[width=\textwidth]{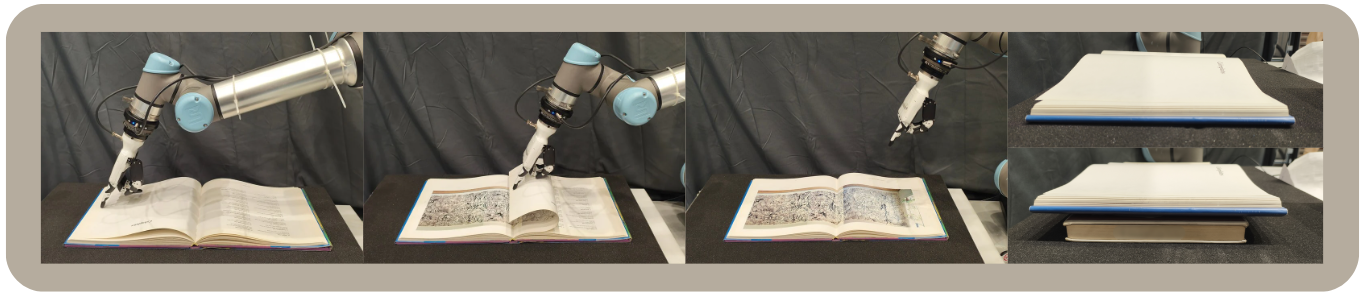} % Replace with your image file
        \subcaption{Page-Flipping: The last figure shows two different height settings. The robot needs to learn to judge the timing of page flipping based on contact force.}
        \label{fig:IL_flip}
        \vspace{2mm}
        \includegraphics[width=\textwidth]{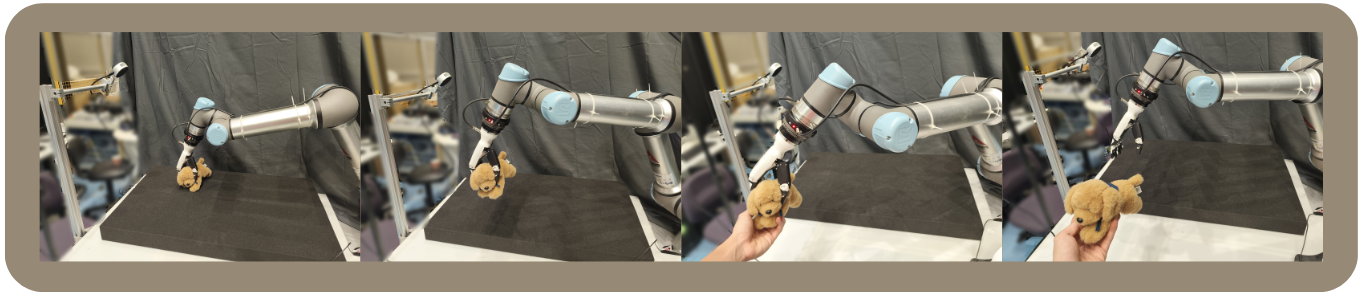} % Replace with your image file
        \subcaption{Hand-Over: The robot grasp a toy dog within a 20$\times$20 cm workspace, moved it to a designated position outside the camera's field of view, and released it upon perceiving a force from receiver.}
    \end{minipage}
    \caption{Imitation Learning Experiment}
\end{figure*}
\begin{figure}[h]
    \centering
    \begin{minipage}{0.48\textwidth}
        \centering
        \includegraphics[width=\textwidth]{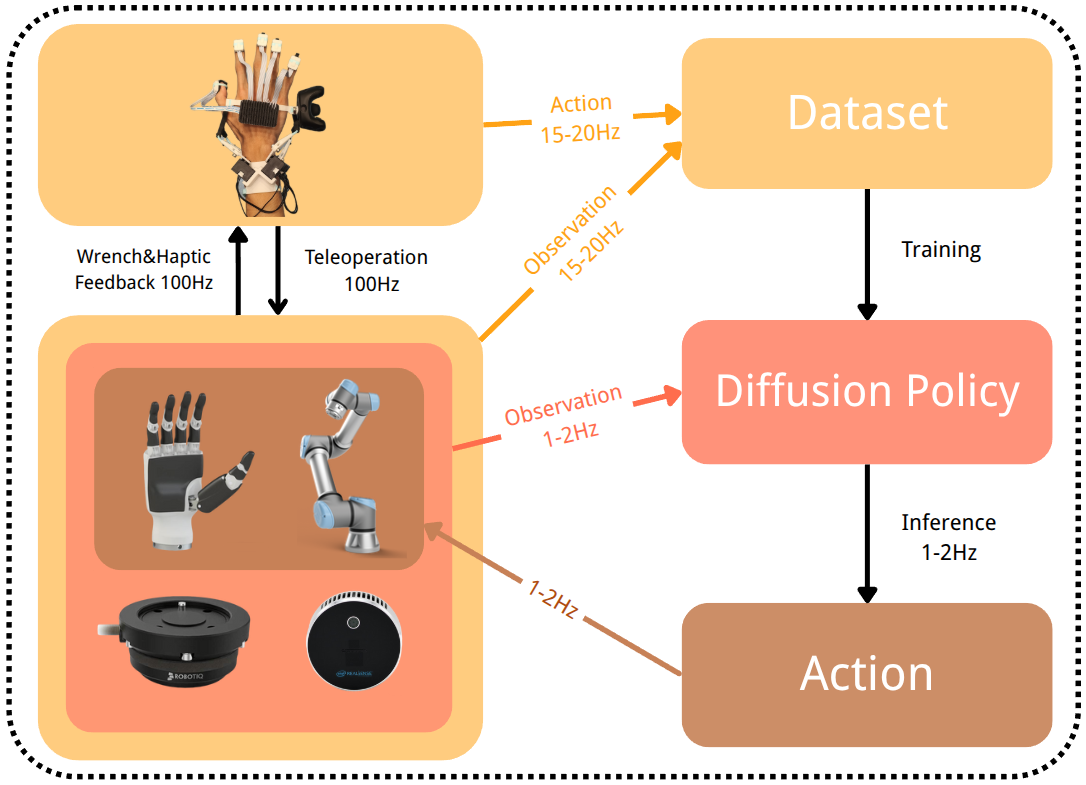} % Replace with your image file
        \caption{Data Flow in Imitation Learning Experiments}
        \label{fig:IL}
    \end{minipage}
\end{figure}
\subsection{Imitation Learning}
We conducted two experiments to evaluate the Glovity system for imitation learning in contact-rich and human-robot interaction tasks, using the IDP3 framework~\cite{ze2025generalizablehumanoidmanipulation3d}. Fig~\ref{fig:IL} shows the data flow in this experiments. An Intel RealSense L515 camera provided RGB input at 640$\times$480, resized to 224$\times$224, the policy integrated RGB images processed via a pretrained R3M encoder and wrench data (6D force-torque vector) as low-dimensional observations, alongside 12D agent pose (6D for UR5e and 6D for Inpire Hand). 
\subsubsection{Page-Flipping Task}
The first experiment assessed the Glovity system's ability to capture high-fidelity wrench data for training policies in contact-rich tasks requiring both force regulation and force-based decision-making. In this setup, a picture album (300$\times$300 mm) was placed on a flat table, with a 6 cm thick hard sponge pad placed between the table and the album to enhance the smoothness of force variations. The first side of the album was randomly elevated to a height between 0 and 5 cm using supports. The paper material and page length required the policy to press with sufficient force and slide backward 6 - 8 cm to initiate page lift, followed by leveraging the page's elasticity to generate friction between the fingertip and page for completing the flipping action. We collected 30 demonstrations, each lasting 150 timesteps (approximately 13s), and trained two policies for 300 epochs: one with wrench data (6D force-torque) as part of the observation space and one without wrench data. Success rate, defined as a complete page flipping without slippage or excessive force causing the robotic arm to stop protectively, was the primary metric.

\textbf{Results:} For book elevations ranging from 0 to 5 cm, the DP-R3M policy augmented with wrench data achieved a success rate of 80\% (16/20 trials), demonstrating robust adaptation to varying book elevations by using force feedback to gauge contact stability before initiating the flip. Ablation studies revealed that removing wrench data from the observations degraded performance to 45\% (9/20 trials), as the policy failed to dynamically adjust the rubbing force and timing, often resulting in incomplete flips or excessive force. This underscores the effectiveness of wrench data in policy learning in enabling adaptive, force-aware behaviors in unstructured environments and the successful utilization of wrench data collected via the Glovity system for effective policy training.

\subsubsection{Hand-Over}

The second experiment explored force-based imitation learning in human-robot interaction. We collected 50 demonstrations, each spanning 200 timesteps (approximately 17s), where the operator grasped a toy dog within a 20$\times$20 cm workspace, moved it to a designated position outside the camera's field of view (50 cm away), and released it upon perceiving a distinct force from another human via wrench and haptic feedback. Two policies were trained for 300 epochs: one with wrench data and one without. Success was defined as the robot releasing the toy only after a pulling force was applied to the toy by a human.

\textbf{Results:} The wrench-augmented policy achieved a 75\% success rate (15/20 trials), reliably holding the toy at the designated position until a human applied sufficient force, then releasing it and returning to the initial pose. Of the five failures, three were due to the robot failing to grasp the toy dog, and two were due to insufficient grip strength, causing the robot to not return to the initial pose after the human removed the toy. Without wrench data, the policy prematurely released the toy upon reaching the designated position, failing to wait for human intervention. However, qualitative analysis revealed that the wrench-augmented policy exhibited slightly increased jitter in the robot's motion, likely due to the diffusion model's iterative inference process exacerbating discontinuities in actions, which amplified noise in the force-torque sensor readings. This issue could potentially be mitigated in future work through optimized data preprocessing or refined model architectures to enhance action continuity.

\section{DISCUSSION\&CONCLUSION}
Glovity exhibits several limitations that warrant further exploration. First, the precision of the wrench feedback system in force rendering remains sub-optimal while the effective feedback area is limited, potentially constraining its utility in tasks demanding precise force interactions, such as microsurgery or fragile object handling. Second, the haptic glove is currently restricted to six degrees-of-freedom tracking, making it challenging to perform complex hand manipulation tasks. Third, our integration of wrench signals into imitation learning frameworks, while effective for basic contact-rich tasks, employs a straightforward observational approach that may not generalize to highly complex, long-horizon manipulations involving variable dynamics.

In conclusion, Glovity represents a significant advancement in cost-effective, open-source teleoperation for contact-rich dexterous manipulation. By integrating spatial wrench feedback with a haptic glove, it facilitates intuitive force-torque and tactile interactions, enhancing data collection and imitation learning. Looking forward, Glovity’s intuitive feedback can be leveraged to advance imitation learning algorithms, allowing models to learn from operators' force-feedback-guided action adjustments, thereby enabling robots to dynamically adapt their movements based on force cues and achieve more complex, long-horizon contact-rich tasks. Furthermore, this force and haptic feedback can facilitate human-in-the-loop learning, empowering humans to correct robotic actions in real-time based on intuitive sensory cues. Additionally, Glovity's ultra-low latency positions it as a promising tool for high-dynamic scenarios, such as locomotion and rapid manipulation. By open-sourcing all hardware and software, we invite the research community to build upon this foundation, addressing limitations like force rendering intuitiveness and expanding degrees-of-freedom, to unlock new frontiers in adaptive, collaborative, and autonomous robotic systems.

 \bibliographystyle{IEEEtran}
 \bibliography{IEEEabrv, reference}

\end{document}